\documentclass[lettersize,journal]{IEEEtran}
\usepackage{amsmath,amsfonts}
\usepackage{algpseudocode}
\usepackage{algorithm}
\usepackage{array}
\usepackage{subcaption}
\usepackage{textcomp}
\usepackage{stfloats}
\usepackage{url}
\usepackage{verbatim}
\usepackage{graphicx}
\usepackage{cite}
\usepackage{hyperref}
\usepackage{adjustbox} % For tables with specified width
\usepackage{booktabs}
\usepackage{multirow}
\usepackage[table,xcdraw]{xcolor}
\usepackage{xcolor}
\usepackage{pifont}
\usepackage{placeins} % Load the placeins package for \FloatBarrier
\usepackage{cite}
\usepackage{tikz}
\usetikzlibrary{shapes.multipart}
\usetikzlibrary{shapes,arrows}
\usepackage{amsmath}
\usepackage{bm}
\begin{document}

\title{TransRAD: Retentive Vision Transformer for Enhanced Radar Object Detection}

\author{Lei Cheng,~\IEEEmembership{Graduate Student Member,~IEEE}, and Siyang Cao,~\IEEEmembership{Senior Member,~IEEE}
        % <-this % stops a space
\thanks{The authors are with the Department of Electrical and Computer Engineering, The University of Arizona, Tucson, AZ 85721 USA (e-mail: leicheng@arizona.edu; caos@arizona.edu)}% <-this % stops a space
}

\markboth{Journal of \LaTeX\ Class Files,~Vol.X, No.X, X}%
{Shell \MakeLowercase{\textit{et al.}}: A Sample Article Using IEEEtran.cls for IEEE Journals}


\maketitle

\begin{abstract}
Despite significant advancements in environment perception capabilities for autonomous driving and intelligent robotics, cameras and LiDARs remain notoriously unreliable in low-light conditions and adverse weather, which limits their effectiveness. Radar serves as a reliable and low-cost sensor that can effectively complement these limitations. However, radar-based object detection has been underexplored due to the inherent weaknesses of radar data, such as low resolution, high noise, and lack of visual information.
In this paper, we present TransRAD, a novel 3D radar object detection model designed to address these challenges by leveraging the Retentive Vision Transformer (RMT) to more effectively learn features from information-dense radar Range-Azimuth-Doppler (RAD) data. Our approach leverages the Retentive Manhattan Self-Attention (MaSA) mechanism provided by RMT to incorporate explicit spatial priors, thereby enabling more accurate alignment with the spatial saliency characteristics of radar targets in RAD data and achieving precise 3D radar detection across Range-Azimuth-Doppler dimensions. Furthermore, we propose Location-Aware NMS to effectively mitigate the common issue of duplicate bounding boxes in deep radar object detection.
The experimental results demonstrate that TransRAD outperforms state-of-the-art methods in both 2D and 3D radar detection tasks, achieving higher accuracy, faster inference speed, and reduced computational complexity. Code is available at \href{https://github.com/radar-lab/TransRAD}{https://github.com/radar-lab/TransRAD}
\end{abstract}


\section{Introduction}
\IEEEPARstart{A}{utonomous} driving and intelligent robotics, both heralded as revolutionary technologies, are steadily becoming a reality \cite{cheng20233d}. Autonomous driving, in particular, holds the promise of significantly reducing human error and enhancing traffic efficiency \cite{chougule2023comprehensive,leimot}, with real-world implementations already underway. However, as these technologies are increasingly deployed, reports of accidents and their limitations have surfaced, casting a shadow over their widespread adoption. Most of these incidents can be traced back to perception errors—specifically, the failure to detect or the incorrect identification of objects. Currently, the perception systems in autonomous vehicles are largely dependent on optical sensors, such as cameras and LiDAR, due to their high resolution and the significant progress made in computer vision research \cite{sengupta2022robust}. Nonetheless, both cameras and LiDAR have unavoidable drawbacks, especially under adverse weather and lighting conditions, including nighttime, glaring sunlight, snow, rain, or fog \cite{gao2020ramp}. In contrast, radar sensors exhibit robustness under these challenging conditions and provide accurate range and velocity information, making them a valuable complement to optical sensors \cite{paek2022k,cheng2023online}. Indeed, numerous studies have shown that integrating radar with optical sensors can significantly improve the reliability and robustness of environmental perception by providing redundancy at the sensor level\cite{sengupta2022robust, liu2021robust, bai2021robust, ouaknine2021multi,decourt2022darod,leimot}. Furthermore, the affordability of radar sensors enhances their appeal as essential components in autonomous driving systems.

Despite these advantages, the exploration of radar-only object detection (especially those based on deep learning) has been relatively slow and limited compared to the extensive research on optical sensors \cite{zhang2021raddet,dalbah2023radarformer,rebut2022raw,kim2022deep,jiang2022t}. This can largely be attributed to the low resolution and high noise levels inherent in radar data \cite{zhang2021raddet,kosuge2022mmwave}, making it exceptionally challenging to extract meaningful features. However, the rapid advancements in deep learning have helped to some extent in overcoming these challenges. Deep learning applications to radar data can be categorized into two main approaches based on the type of radar data representation used. The first approach involves point cloud-based methods, which can leverage well-established techniques from LiDAR point cloud processing. However, radar point clouds are much sparser than LiDAR point clouds, making it challenging to learn meaningful features \cite{wang2021rodnet, zou2023transrss,giroux2023t,brodeski2019deep}. As a result, this method is more commonly applied in sensor fusion techniques that involve radar. The second approach is based on RAD cube data. The RAD data, being a lower-level representation of radar data, preserves more detailed radar detection information \cite{ouaknine2021multi,wang2021rodnet,zhuang2023effective} and resembles image data \cite{dalbah2023radarformer}. This makes it more suitable for deep learning to extract useful radar features, and thus, RAD-based methods are emerging as the mainstream approach for radar-only object detection.

However, most current RAD-based methods rely on convolutional neural networks (CNNs) \cite{kim2022deep,major2019vehicle,wang2021rodnet,wang2021rodnet2,gao2020ramp,ouaknine2021multi,huang2022yolo,zhang2021raddet,decourt2022darod,ju2021danet,kothari2023object}, which often struggle to capture global context information and long-range semantic dependencies \cite{dalbah2024transradar, zou2023transrss,jiang2022t}. Among these, 3D CNNs are the most commonly used due to their ability to process 3D spatial data \cite{major2019vehicle,wang2021rodnet,wang2021rodnet2,gao2020ramp,ouaknine2021multi,ju2021danet}, but they are notoriously computationally expensive. In contrast, Transformers excel at capturing global context and long-range dependencies through their self-attention mechanisms \cite{islam2022recent}, which can process complex radar data more effectively. Despite these advantages, only a few RAD-based methods have utilized transformer networks \cite{jiang2022t,zhuang2023effective,dalbah2023radarformer}, and these are typically combined with 3D CNNs. This combination, while powerful, significantly increases computational complexity due to the inherent complexity of both models.
Moreover, none of these methods adequately address the unique characteristics of radar object detection. Radar objects typically occupy irregularly shaped and small regions in the RAD spectrum, making them akin to small object detection. Additionally, radar targets exhibit a distinctive spatial distribution in RAD data, where the central region of the target area exhibits the highest intensity, which gradually diminishes towards the edges, as shown in Fig. \ref{rod}.

To bridge these research gaps, we propose TransRAD, a retentive vision transformer-based deep radar detector that uses RAD data as input, strategically developed to leverage the strengths of Transformers and circumvent the computationally intensive 3D CNNs. Our model, as shown in Fig. \ref{transrad}, is structured around three key components: the Backbone, Neck, and Head, each designed to address specific challenges in radar object detection.
For the Backbone, we employ an RMT \cite{fan2024rmt}, which utilizes explicit spatial decay priors (i.e., MaSA). This design enables the model to perceive global information while assigning varying levels of attention to tokens at different distances \cite{fan2024rmt}, effectively aligning with the spatial characteristics of radar targets in RAD data. Additionally, we incorporate multi-scale feature extraction to leverage the multi-scale semantic information from radar data, thereby improving detection accuracy across different object sizes. 
For the Neck, we use a feature pyramid network (FPN) \cite{lin2017feature}, enhancing the semantic richness of the larger feature layers, which is particularly advantageous for detecting small objects. 
For the Head, inspired by YOLOv8 \cite{Yolov8}, we adopt anchor-free detection heads to better handle the small and irregularly shaped radar objects \cite{zhang2019freeanchor,duan2020corner,ouyang2023anchor}, completely avoiding the complicated computations related to anchor boxes \cite{tian2019fcos}. To further refine the detection process, unlike most existing radar object detection methods that use coupled heads, we implement the decoupled heads to separately perform objectness prediction, classification, bounding box regression, and Doppler-specific regression, thus reducing interference between these tasks and enhancing the model’s overall performance \cite{ge2021yolox}. Moreover, unlike most current methods that generate 2D bounding boxes, our approach aims to produce 3D bounding boxes. To maintain strong 2D detection performance while advancing to 3D, we incorporate both 3D bounding box loss and 2D bounding box loss in the loss function. This strategy ensures that our model maintains high accuracy in 2D detection (in the RA and RD planes) while also advancing the capabilities of 3D object detection (in the RAD cube), effectively addressing the unique challenges of interpreting radar data. We also explicitly include a center loss to address the critical requirement for precise center detection in radar object detection. Lastly, to address the inherent weakness of radar data in classification tasks, we introduce a two-stage NMS approach, wherein classification-based NMS is followed by Location-Aware Non-Maximum Suppression (LA-NMS), thereby capitalizing on radar object detection's higher localization precision.
The main contributions of our work can be summarized as follows:
\begin{enumerate}
  \item  We propose TransRAD, an RMT-based deep radar detector. It uses explicit spatial decay priors to match the distinctive spatial distribution of radar targets in RAD data, where the central region exhibits the highest intensity, gradually diminishing towards the edges. Additionally, it leverages the transformer's ability to capture global context while avoiding the complexity of 3D CNNs.
  \item  Our model employs an FPN Neck and four task-specific anchor-free decoupled detection heads. This design effectively addresses the small and irregular shapes of targets in radar object detection to improve detection accuracy while also preventing overfitting due to the limited radar training data.  
  \item  TransRAD generates 3D bounding boxes, allowing for the determination of the distance, angle, and velocity of radar objects in a single step. It further integrates 3D and 2D bounding box losses, as well as center loss, ensuring robust 2D and center detection performance while advancing 3D object detection capabilities.  
  \item  We propose an LA-NMS strategy that is applied after the classification-based NMS. This enables us to capitalize on radar object detection's high location accuracy to compensate for its classification limitations. Our experiments reveal that this approach effectively eliminates overlapping bounding boxes with different classes, which is particularly significant given the low probability of radar target overlap in reality.
\end{enumerate}

The rest of this paper is organized as follows: Section II presents some background information and a review of related works. Section III describes our proposed method in detail. Section IV provides experimental results and analysis, followed by conclusions in Section V.

\section{Background and Related Works}
\subsection{Radar Object Detection}
\label{sec:Radar Object Detection}
Object detection is a technique for locating objects and identifying their classes in input data. Radar object detection specifically uses radar data as input for object detection tasks. Currently, object detection is commonly based on deep learning due to its extraordinary accuracy. Given the complexity of radar data—characterized by low resolution and high noise—radar object detection particularly relies on deep learning.
Radar data typically comes in three forms: analog-to-digital (ADC) data, point cloud data, and RAD cube data. ADC data is generally considered the raw radar data, often structured as a 3D Sample-Chirp-Antenna cube \cite{yao2023radar}. This format has an extremely large data size and is difficult to interpret for downstream tasks \cite{zou2023transrss}. By applying a three-dimensional (3D) Fast Fourier Transform (FFT), ADC data can be transformed into RAD cube data. Using Constant False Alarm Rate (CFAR)-based peak detection processing, RAD cube can further be converted into the radar point cloud. Point cloud data contains limited information, making it challenging to extract useful semantics. In contrast, RAD data not only preserves most of the information available in the raw ADC data but also provides an image-like data representation \cite{patel2019deep}. This makes it well-suited for vision-based deep learning models  \cite{patel2019deep}, enabling more effective radar object detection. Therefore, our work is based on RAD data.

\begin{figure}[htbp]
	\centering
	\includegraphics[width=0.499\textwidth]{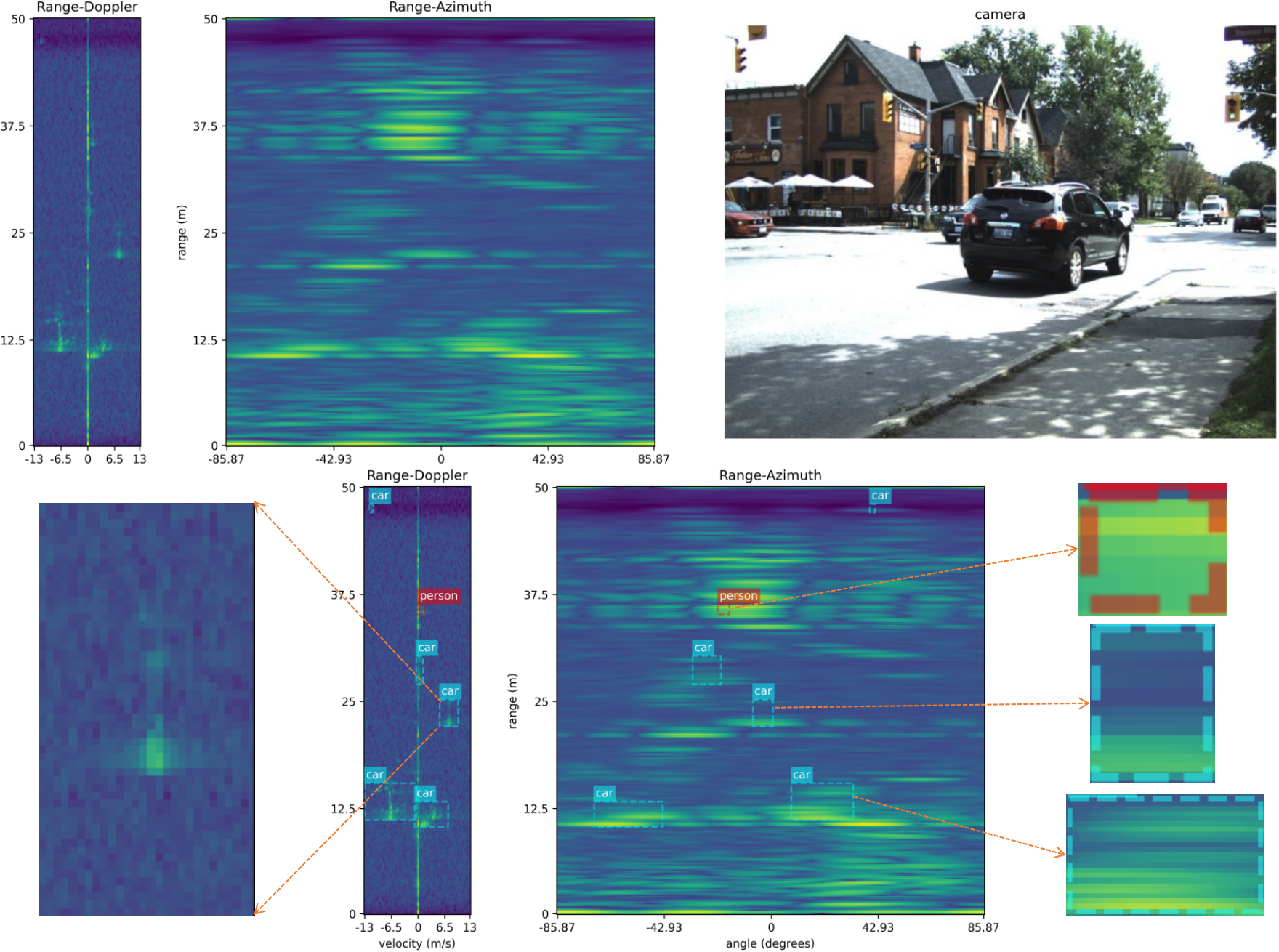}
	\caption{Unique aspects of radar object detection.}
	\label{rod}
\end{figure}

Radar object detection, as we have found, significantly differs from general image object detection in several key aspects, illustrated in Fig. \ref{rod}:

\begin{enumerate}
    \item \textbf{Irregular Shapes and Lack of Visual Correlation}: Radar targets in RAD data typically have irregular shapes, with non-uniform and diverse outlines. Additionally, the size and shape of radar targets are not directly correlated with the actual size and appearance of real-world objects, lacking the visual information present in traditional images.
    
    \item \textbf{Small Target Areas}: Radar targets often occupy small areas. According to the definition of small objects—those occupying an area less than or equal to 1024 pixels \cite{cheng2023towards}—most radar targets fall into the category of small or even tiny objects. Detecting small objects is inherently challenging due to their limited size and the generic feature extraction paradigm, which usually down-samples feature maps to reduce spatial redundancy and learn high-dimensional features \cite{cheng2023towards}. This process inevitably diminishes the representation of small objects, leading to low-quality feature representation.     
    
    \item \textbf{Saliency Detection}: Radar object detection can be considered a type of saliency detection \cite{wang2021salient}. The target in the radar RAD data exhibits a high-intensity core that gradually diminishes towards the outer edges, standing out distinctly against the relatively uniform background.
    
    \item \textbf{Unreliable Classification}: Radar data's low resolution and high noise, combined with the absence of detailed visual cues like those in image targets, make it more challenging to achieve reliable classification in radar object detection \cite{kosuge2022mmwave}. Radar is typically used as a range-only sensor, prioritizing location precision over classification accuracy.   
    
    \item \textbf{3D Nature of RAD Data}: Radar RAD data is inherently 3D, differing from image data where the three RGB channels are strongly correlated. In contrast, the three dimensions of RAD data independently represent range, azimuth, and Doppler, each conveying different information. Therefore, for radar object detection, generating a 3D bounding box to capture information across all three dimensions is more meaningful.
\end{enumerate}

Our work is built upon these unique aspects of radar object detection, aiming to fully address and leverage these characteristics.

\subsection{CNN-based Radar Object Detection}
CNNs are widely used in radar object detection due to their superior performance in visual object detection tasks. The authors in \cite{gao2020ramp} introduced RAMP-CNN, which slices the 3D RAD cube into three 2D images—Range-Azimuth (RA), Range-Doppler (RD), and DA heatmaps. Three parallel 3D CNN autoencoders then process these heatmaps to generate distinct feature bases, which are subsequently fused to make object recognition decisions.
The approach in \cite{major2019vehicle} is similar to RAMP-CNN, where the 3D RAD cube is also sliced into three 2D images, each processed through multiple convolutional layers to obtain individual feature maps. These feature maps are then concatenated and further processed by 3D convolutional layers, followed by an FPN for multi-scale feature fusion.
In \cite{wang2021rodnet} and \cite{wang2021rodnet2}, the authors proposed RODNet, which is based on 3D convolutional autoencoder networks, and utilized a camera-radar fusion cross-modal supervision framework to train it.
The authors in \cite{ju2021danet} presented the Dimensional Apart Network (DANet), which reduces the computational load of RODNet using a Dimension Apart Module (DAM). However, it still relies on 3D CNNs.

All these methods employ 3D CNNs, but the computational complexity of 3D CNNs is prohibitive. For instance, training RODNet can take up to a week \cite{ju2021danet}, making the use of 3D CNNs impractical. Additionally, these methods focus on point target detection rather than generating bounding boxes.
Some other works have explored the use of 2D CNNs for radar object detection. The authors in \cite{kim2022deep} and \cite{huang2022yolo} both employed YOLO-based deep learning models to detect 2D bounding boxes on RA maps. In \cite{decourt2022darod}, a lightweight DAROD detector was proposed for 2D bounding box object detection in RD spectra.
The authors in \cite{zhang2021raddet} proposed a ResNet-based RadarResNet that uses RAD data as input to generate 3D bounding box detection. This approach showed that 2D CNNs could effectively process 3D radar data to achieve 3D detection results. However, it uses an anchor-based method, which is less effective than anchor-free detection when dealing with irregularly shaped radar targets.
\begin{figure*}[htbp]
	\centering
	\includegraphics[width=0.99\textwidth]{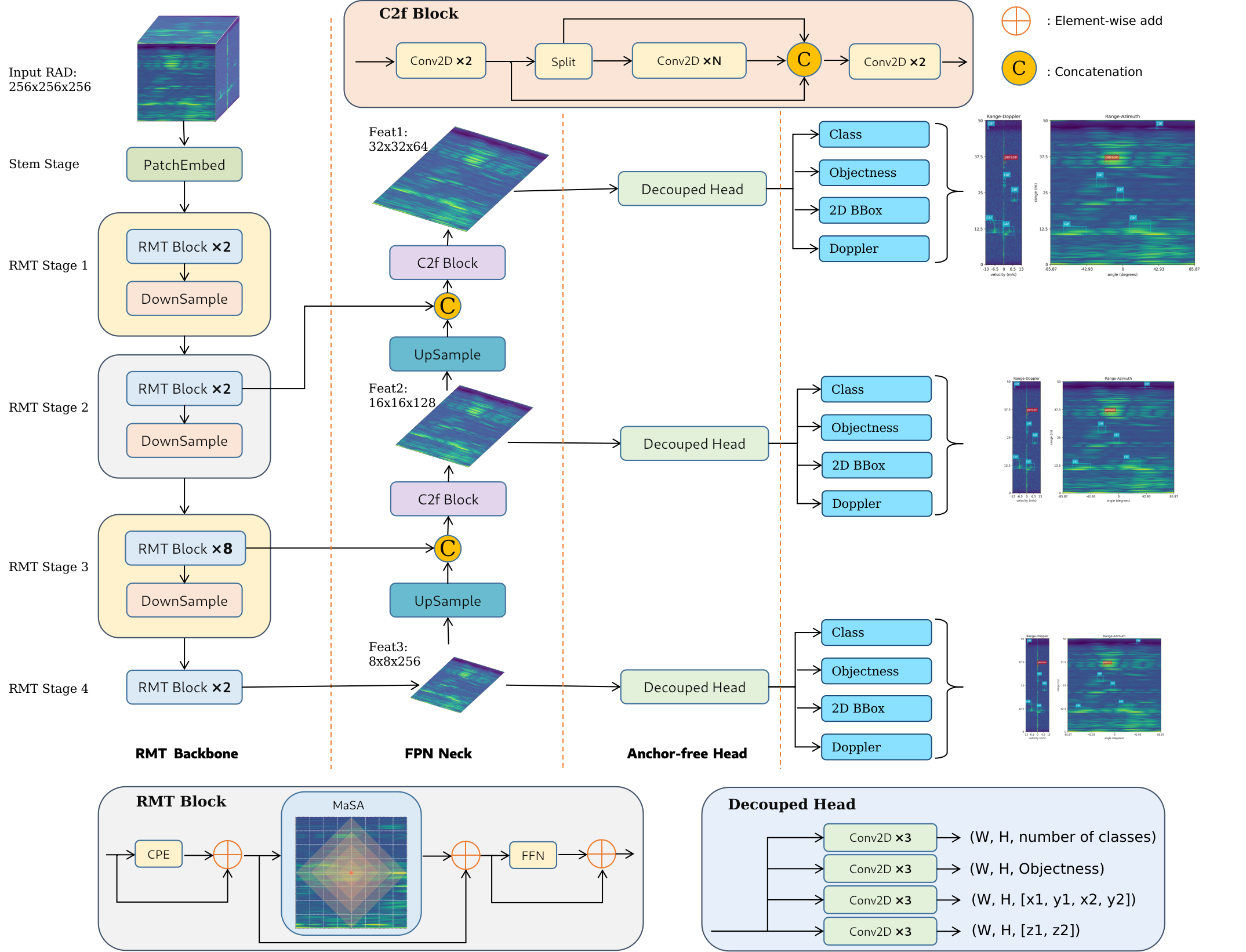}
	\caption{Overall architecture of TransRAD.}
	\label{transrad}
\end{figure*}

\subsection{Transformer-based Radar Object Detection}
Transformers have become the dominant model in natural language processing and are increasingly showing potential to replace CNNs in visual processing tasks \cite{yang2022transformers}. Compared to CNNs, Transformers offer advantages such as better capability to capture long-range dependencies and global context through their self-attention mechanisms \cite{islam2022recent}, and improved scalability with larger datasets \cite{giroux2023t}. Therefore, Transformer-based radar object detection holds significant promise.
The authors in \cite{dalbah2023radarformer} proposed RadarFormer, a hybrid model combining 3D CNNs and MaXViT (Multi-Axis Vision Transformer). This model aims to reduce the total computational complexity and training/inference times of the RODNet model.
In \cite{jiang2022t}, the authors introduced T-RODNet, which uses a 3D CNN combined with a 3D Swin Transformer. It integrates the DAM and T-window-multi-head self-attention (T-W-MSA)/shifted window multi-head self-attention (SW-MSA) modules for efficient multi-scale feature fusion.
Similarly, the authors in \cite{zhuang2023effective} developed SS-RODNet by combining 3D CNNs with a 3D Swin Transformer and incorporating a masked image modeling self-supervision method. This approach aims to reduce the complexity of T-RODNet while enhancing the network’s performance with a limited dataset.
Additionally, \cite{dalbah2024transradar} and \cite{zou2023transrss} utilized transformers for radar semantic segmentation tasks, not radar object detection. While \cite{dalbah2024transradar} still combines 3D CNNs with Transformers, \cite{zou2023transrss} uses RA and RD views as input data instead of the RAD cube.

Despite these studies demonstrating the potential of Transformers to improve accuracy in radar object detection, they almost always combine Transformers with 3D CNNs. Both Transformers and 3D CNNs are highly complex models, and their combined use significantly increases computational complexity.
To the best of our knowledge, there are almost no radar object detection models that are purely transformer-based or that combine 2D CNNs with Transformers. Moreover, none of these existing models, whether CNN-based or transformer-based, have fully considered the unique characteristics of radar object detection that we outlined in Sec. \ref{sec:Radar Object Detection}.

\section{Proposed Methodology}

Our approach leverages an RMT as the Backbone and 2D CNNs for the Neck and Head, aiming to avoid the high computational complexity of 3D CNNs while fully considering the unique aspects of radar object detection to improve its performance. In this section, we begin by explaining the suitability of MaSA for radar object detection. Then, we detail the proposed TransRAD model and loss functions, demonstrating how they address the specific challenges inherent in radar object detection as discussed in Sec. \ref{sec:Radar Object Detection}.
Specifically, the Backbone of TransRAD, through the incorporation of explicit spatial priors, is tailored to manage the spatial saliency decay characteristics of radar targets. The Neck utilizes multi-scale features to handle the challenge of small radar targets. The anchor-free decoupled heads are designed to cope with the irregular shapes of radar targets, while the RA 2D detection head and Doppler head enable the simultaneous determination of distance, angle, and velocity. The detailed architecture of TransRAD is shown in Fig. \ref{transrad}.
In the loss function, we have integrated a 2D bounding box loss to reflect the common practice of presenting radar detections in 2D bird's-eye views. Additionally, the inclusion of center loss enhances the accuracy of radar target localization and mitigates the issue of low tolerance for bounding box perturbation in small objects \cite{cheng2023towards}. Lastly, a Location-Aware NMS approach was implemented to address the issue of duplicate bounding boxes from different classes, caused by unreliable classification in radar object detection.

\subsection{Retentive Manhattan Self-Attention}
\begin{figure}[htbp]
	\centering
	\includegraphics[width=0.499\textwidth]{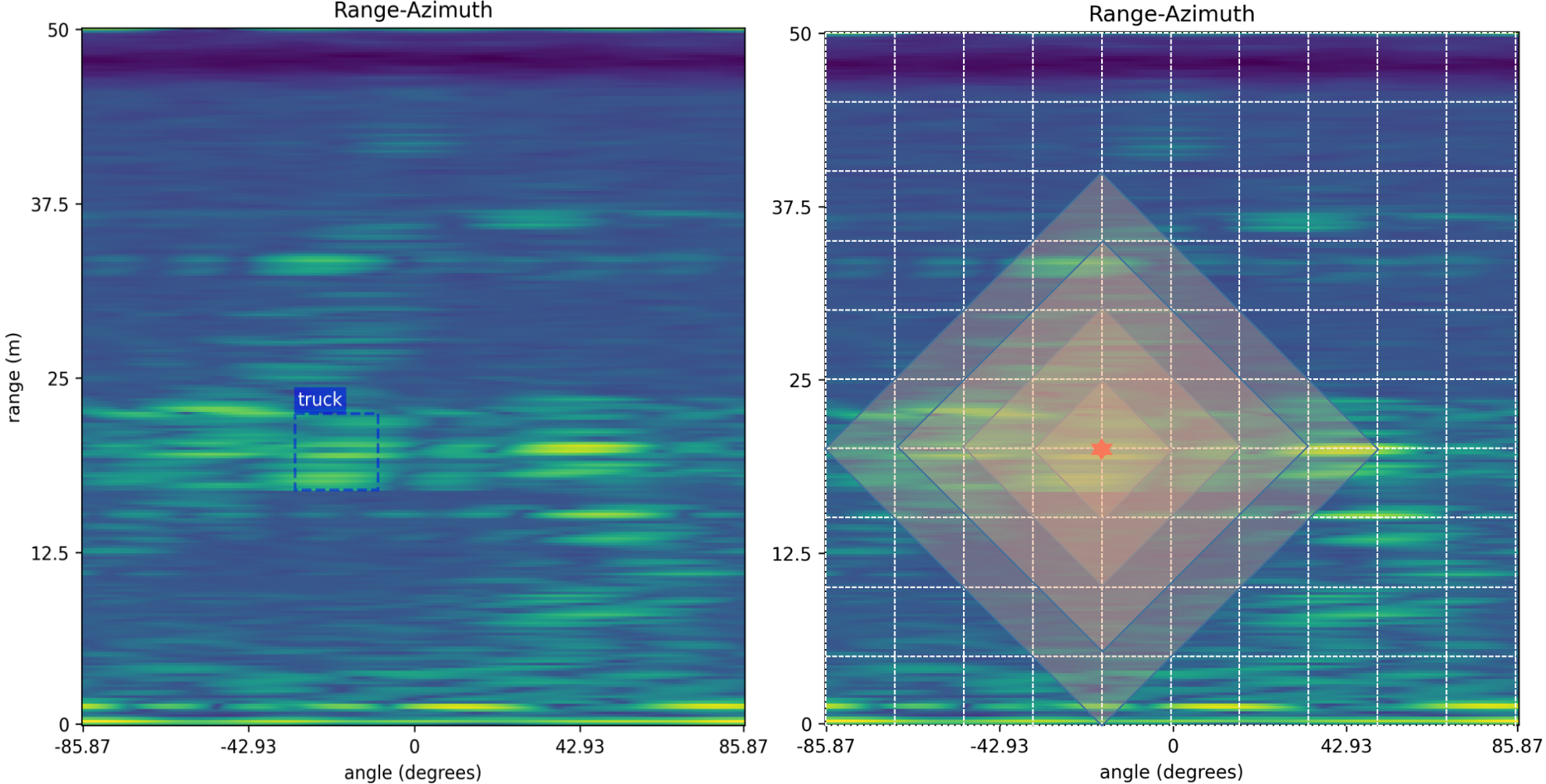}
	\caption{Explicit spatial prior in MaSA: attention diminishes with increasing Manhattan distance from the center.}
	\label{masa}
\end{figure}

Radar object detection has a most notable characteristic: the target in the radar RAD data exhibits a high-intensity core that gradually diminishes towards the outer edges. This makes the target stand out clearly against a relatively uniform background. Effective radar object detection, therefore, requires a global perception that focuses more on the object's center and its immediate surroundings. However, the original self-attention mechanism in Transformers fails to achieve this, as it applies uniform attention to the entire frame of data. This not only reduces the precision of radar object detection but also incurs unnecessary computational costs \cite{fan2024rmt}.
In contrast, MaSA introduces an explicit spatial prior based on the Manhattan distance, allowing it to capture global context while paying more attention to the object’s center and surrounding areas, as shown in Fig. \ref{masa}.

MaSA is inspired by the retention mechanism introduced in the Retentive Network (RetNet) \cite{sun2023retentive}. The retention mechanism uses a unidirectional and one-dimensional temporal decay for sequence modeling, achieving better inference efficiency, improved training parallelization, and competitive performance compared to Transformers \cite{sun2023retentive}.  This mechanism can be defined as:
\begin{equation}
o_n = \sum_{m=1}^{n} \gamma^{n-m} \left( Q_n e^{in\theta} \right) \left( K_m e^{im\theta} \right)^\dagger v_m \ ,
\end{equation}
and its parallel representation as:
\begin{equation}
\begin{split}
Q = (X W_Q) \odot \Theta, \quad 
K &= (X W_K) \odot \overline{\Theta}, \quad 
V = X W_V , \\
\Theta_n = e^{in\theta}, \quad 
D_{nm} &= \begin{cases} 
\gamma^{n-m}, &n \geq m \\
0, &n < m 
\end{cases}  \ , \\
\text{Retention}(X) &= (Q K^T \odot D) V \ ,
\end{split}
\end{equation}
where \( o \) is the attention scores output, \( X \) is the input sequence, \( m \) and \( n \) are token indices, \( \gamma \) is a decay ratio scalar less than 1, \( Q \) (Query), \( K \) (Key), and \( V \) (Value) are projection vectors derived from the input sequence, \( W_Q \), \( W_K \), and \( W_V \) are learnable matrices, \( \dagger \) is the conjugate transpose, \( \odot \) is the Hadamard product, \( \Theta \) is the complex exponential function, \( \overline{\Theta} \) is the complex conjugate of \( \Theta \), and \( D \in \mathbb{R}^{|x| \times |x|} \) is a temporal decay matrix which combines causal masking and exponential decay along relative distance.
MaSA extends this retention mechanism into a bidirectional and two-dimensional spatial decay format. First, the retention is expanded into a bidirectional form:
\begin{equation}
D_{nm}^{Bi} = \gamma^{|n-m|} = \begin{cases} 
\gamma^{n-m}, &n \geq m \\
\gamma^{m-n}, &n < m 
\end{cases}
\end{equation}
Then, it is further extended to two dimensions:
\begin{equation}
D_{nm}^{Bi,2d} = \gamma^{|x_n - x_m| + |y_n - y_m|} \ ,
\end{equation}
where \( (x_n, y_n) \) and \( (x_m, y_m) \) are the two-dimensional coordinate for the \( n \)-th token and the \( m \)-th token, respectively, and \( |x_n - x_m| + |y_n - y_m| \) represents the Manhattan distance between the \( n \)-th token and the \( m \)-th token. Finally, Softmax is employed to introduce nonlinearity. The complete form of MaSA now is expressed as:
\begin{equation}
\text{MaSA}(X) = \left( \text{Softmax}(Q K^T) \odot D^{Bi,2d} \right) V .
\end{equation}

Clearly, according to the space decay matrix \( D^{Bi,2d} \), the farther the surrounding tokens are, the greater the degree of decay in their attention scores. This feature allows it to assign different levels of attention to tokens at varying distances \cite{fan2024rmt}. Ultimately, it enables the perception of global information while giving more attention to the area immediately surrounding the target token. This perfectly aligns with the characteristics of radar objects in RAD data.

\subsection{Backbone Based On RMT}
\label{sec:backbone}
Our proposed radar object detection model, TransRAD, employs RMT \cite{fan2024rmt} as the Backbone to fully leverage its MaSA mechanism, which explicitly assigns different levels of attention based on spatial distances. RMT, a recently introduced vision transformer backbone, uses a distance-dependent spatial decay matrix to provide explicit spatial priors for 2D image data. It has achieved state-of-the-art (SOTA) accuracy in various vision tasks such as image classification and object detection. Additionally, RMT introduces the Decomposed MaSA to address the high computational costs associated with Vanilla Self-Attention, which is used by MaSA for global information modeling. This approach computes attention scores separately for horizontal and vertical directions:
\begin{equation}
\begin{split}
D_{nm}^{Bi,H} = &\gamma^{|y_n - y_m|}, \quad
D_{nm}^{Bi,W} = \gamma^{|x_n - x_m|}, \\
MaSA_H &= \text{Softmax}(Q_H K_H^T) \odot D^{Bi,H}, \\
MaSA_W &= \text{Softmax}(Q_W K_W^T) \odot D^{Bi,W}, \\
\text{MaSA}(X) &= MaSA_H (MaSA_W V)^T.
\end{split}
\end{equation}

Furthermore, RMT uses Depthwise Convolution (DWConv) to implement a Local Context Enhancement (LCE) module. This module addresses the transformer's inherent weakness in local feature extraction (especially compared to CNNs), endowing RMT with improved local perception capabilities. The overall output of the MaSA is computed as follows:
\begin{equation}
\text{MaSA}_{out} = \text{MaSA}(X) + \text{LCE}(V).
\end{equation}
%Choosing RMT as the backbone for TransRAD was strategic. 

The decision to use RMT as the Backbone for TransRAD was strategic and driven by its dual capabilities: the traditional transformer’s strength in global context awareness and its enhanced ability to focus on local regions. This combination is particularly well-suited for radar object detection tasks, where both global understanding and local details are crucial. In fact, providing global context has numerous benefits for radar object detection. It increases localization precision \cite{ren2020salient}, helps in detecting small objects \cite{liu2021survey,wu2022gcwnet}, and improves the overall completeness of object detection \cite{chen2020global}. This is where RMT excels over CNNs.

RMT was initially designed for image object detection, but radar object detection has unique requirements. Firstly, although radar data might appear complex, it contains less diverse information compared to camera images, which include varied textures, orientations, geometry, lighting, and are generally larger \cite{decourt2022darod} (often exceeding 416 × 416). Radar data, despite being noisier, tends to exhibit more similar patterns and shapes with relatively fixed orientations \cite{decourt2022darod} and is typically smaller (usually below 256 × 256).
Moreover, unlike image-based object detection which needs deep networks to enlarge receptive fields, radar targets are smaller and more localized within RAD data \cite{kothari2023object}. Hence, even networks with small receptive fields can perform effectively. 

Given these differences, we here configured RMT as a lighter Backbone to better suit the demands of radar object detection. 
Specifically, the architecture begins with a stem stage based on the Patch Embedding (PatchEmbed) module, composed of four 3 × 3 convolutions, followed by four stages. Each stage, except the third, consists of 2 RMT Blocks, while the third stage is composed of 8 RMT Blocks. Each RMT Block comprises a Conditional Position Encoding (CPE) \cite{chu2021conditional} with 3 × 3 depth-wise convolutions, MaSA, and a Feed Forward Network (FFN).
The output embedding dimensions of these four stages are 32, 64, 128, and 256, respectively. Notably, the first three stages utilize the Decomposed MaSA, while the last stage employs the original MaSA. Patch Merging layers are used between stages, applying 3 × 3 convolutions with a stride of 2 to downsample the feature map. Our Backbone outputs three feature maps from the final three stages, as shown in Fig. \ref{transrad}.

\subsection{Neck Based On FPN}
\label{sec:neck}
Feature Pyramid Network (FPN) has been adopted as TransRAD's Neck to utilize multi-scale features to address the challenge of detecting small radar targets.
Detecting small objects is inherently difficult. The first challenge is that small objects contain limited information and are hard to distinguish from noisy clutter in the background \cite{cheng2023towards}, complicating accurate boundary localization.
Our RMT Backbone has partially addressed this challenge. RMT captures global context through its self-attention mechanism, facilitating faster and more accurate localization of small targets \cite{mottaghi2014role}. By allocating different attention to various parts of the feature maps, it highlights regions valuable for small object detection while suppressing irrelevant background noise.
However, to achieve strong feature extraction and generalization capabilities, networks usually need to go deeper, and our TransRAD is no exception. This depth can cause the limited information of small objects to be nearly wiped out in higher-level features \cite{cheng2023towards}, presenting another challenge.

Multi-level feature fusion has proven effective in mitigating this issue \cite{lin2017feature,wu2022gcwnet}. Deep networks inherently produce multi-level features, each corresponding to different scales. Low-level features retain more details and have higher resolution but contain more noise, whereas high-level features possess stronger semantic information and are more robust to noise, variations, and transformations. Therefore, combining these features will benefit small object detection \cite{liu2021survey}, and FPN exemplifies this approach. FPN uses a top-down pathway and lateral connections to merge low-level features with high-level features, creating a feature pyramid that has rich semantics at all levels \cite{lin2017feature}. Following this practice, we adapt the YOLOv8 Neck to perform only FPN-fashion top-down fusion on the three feature maps output by the TransRAD Backbone, as depicted in Fig. \ref{transrad}, where C2f refers to a faster implementation of the CSP (Cross Stage Partial) \cite{Yolov8} Bottleneck with 2 convolutions module. By doing so, we can harness the strengths of both high-level and low-level features, facilitating coarse radar object localization and boundary refinement, respectively.
Notably, high-level features can be diluted as they pass through the top-down pathway. However, our use of Transformers ensures that each layer inherently contains global semantics, mitigating this dilution effect \cite{chen2020global}.

\subsection{Anchor-Free Decoupled Head}
\label{sec:head}
Anchor-based detection employs pre-defined anchor boxes at multiple scales and aspect ratios as references for object localization \cite{zhang2019freeanchor}, achieving notable performance and becoming a widely used method. However, anchor-based detection comes with several drawbacks. Firstly, achieving optimal detection performance requires clustering analysis to determine a set of optimal anchors prior to training. These clustered anchors are domain-specific and lack generalization \cite{ge2021yolox}. Secondly, the anchor mechanism increases the complexity of detection heads and the number of predictions for each image \cite{ge2021yolox}. Third, fixed-size and fixed-aspect-ratio anchor boxes struggle to handle objects with various geometric layouts, especially those that are eccentric, slender\cite{zhang2019freeanchor}, or irregularly shaped, which is a common characteristic of radar objects. Additionally, only common object sizes and shapes are selected as anchors, leading to the potential omission of objects with peculiar shapes \cite{duan2020corner}.

Anchor-free methods have recently been shown to effectively address these limitations and achieve better performance \cite{tian2019fcos,zhang2020bridging}. Thus, we chose to implement our detection head using an anchor-free approach. Specifically, we adopted the anchor-free method employed in YOLOv8. This method divides each of the three feature maps output by the Neck into an \( S \times S \) grid, where the grid cell containing the center of an object, which is considered a positive sample\cite{feng2021tood}, is responsible for detecting that object \cite{zhang2020bridging}.
By eliminating the usage of anchor boxes, the anchor-free approach not only avoids the extensive computations associated with anchor boxes but also better accommodates the irregular geometric shapes of radar objects(typically eccentric and slender, such as in Fig. \ref{rod}).

Coupled heads use a single head for all prediction tasks, saving parameters and reducing model complexity compared to decoupled heads, where each task has its own head. Almost all existing radar object detection models adopt coupled heads \cite{kim2022deep,major2019vehicle,wang2021rodnet,wang2021rodnet2,gao2020ramp,huang2022yolo,zhang2021raddet,decourt2022darod,ju2021danet,kothari2023object,jiang2022t,zhuang2023effective,dalbah2023radarformer}, likely because these models often employ 3D CNNs, which are already highly complex. Thus, they prefer coupled heads to avoid further increasing model complexity. However, having distinctly different tasks (such as classification and regression \cite{feng2021tood}) share a single coupled head with the same parameters can inevitably cause conflicts \cite{song2020revisiting}. Decoupled heads, on the other hand, prevent this conflict by using separate heads for each task, enabling task-specific predictions\cite{song2020revisiting}.

Given this, and considering our low-complexity Backbone and Neck (without using 3D CNNs), we employ four different heads to perform objectness prediction, classification, RA 2D bounding box regression, and Doppler-specific regression (the maximum and minimum points regression in the Doppler dimension). This separation allows each head to specialize in its task, improving overall performance. For radar object detection, accurate prediction of object presence is particularly important, owing to the inherently noisy nature of the radar data which can lead to frequent false positives and false negatives. Thus, we have a dedicated head for objectness prediction.
We also aim to produce radar 3D detections, which allow us to determine an object's distance, direction, and speed in one step. Typically, 3D object detection requires using 3D convolutions to generate 3D feature maps (excluding the channel dimension) at different scales, which is computationally and memory intensive. With 3D feature maps, an anchor-free approach would involve further dividing them into an \( S \times S \times D \) grid, with the grid cell containing the center of an object as a positive sample. However, this significantly increases the number of grid cells (by a factor of \( D \)) and the associated computations, exacerbating the imbalance between positive and negative samples. For example, assuming \( T \) positive samples, an \( S \times S \) grid has \( S \times S - T \) negative samples, while an \( S \times S \times D \) grid has \( S \times S \times D - T \) negative samples.

To avoid the complexity of 3D CNNs and the extensive computations of 3D feature maps, we treat the Doppler dimension of the RAD data as a channel to generate 2D feature maps in the RA plane. We then use a 2D bounding box regression head to predict the RA 2D bounding box, represented by $[x1, y1, x2, y2]$. With precise RA 2D bounding boxes, predicting the maximum and minimum points in the Doppler dimension, i.e., $[z1, z2]$, becomes simpler. Notably, we assume that our model can learn Doppler-specific information when it uses the complete RAD data as input, as supported by \cite{zhang2021raddet}. 
Since the Doppler-specific regression task differs from the RA 2D bounding box regression task, and to avoid these tasks affecting the accuracy of each other, we use a separate Doppler head for predicting the maximum and minimum points in the Doppler dimension.
Consequently, we can generate 3D bounding boxes, represented as $[x1, y1, z1, x2, y2, z2]$, for radar objects without extensive computations. Fig. \ref{transrad} illustrates the architecture and output shapes of these dedicated heads.

\subsection{Loss Function}
\label{sec:loss}
Once the heads of our model generate the predicted feature maps, the next step is to assign these predictions to the corresponding ground truth labels to compute the loss. This loss is pivotal in supervising the model's training and learning process, guiding the model through iterative updates to achieve higher accuracy and better generalization. Following the practice of YOLOv8, we employ task alignment learning (TAL) \cite{feng2021tood} to allocate the top-K positive samples for each ground truth instance, with setting K = 10 in our case. TAL uses an anchor alignment metric to explicitly measure the degree of task alignment at the anchor level (for anchor-free methods, the anchor is the grid cell). The anchor alignment metric is defined as follows:
\begin{equation}
t = c^{\alpha} \times l^{\beta},
\label{eq:aa_metric}
\end{equation}
where \( c \) and \( l \) denote a classification score and a localization score, respectively. \( \alpha \) and \( \beta \) are used to control the impact of the two tasks in the anchor alignment metric. 
Based on this metric, it dynamically selects high-quality anchors, characterized by both a high classification score and precise localization, to improve detection accuracy and alleviate the misalignment of classification and localization \cite{feng2021tood} that can arise with decoupled heads.

After label assignment, we can now design the loss function to calculate the loss. Various loss functions have been proposed for different tasks, and selecting an inappropriate loss function can undoubtedly hinder the model's ability to learn the desired features \cite{wang2020comprehensive}, leading to decreased performance. Therefore, for our model, we combined four task-specific loss functions to provide more supervisory signals, enabling the model to achieve optimal performance across these tasks. 

Firstly, for the 2D bounding box regression task, we use the Complete-Intersection over Union (CIoU) loss \cite{zheng2021enhancing}. CIoU incorporates IoU loss, normalized central point distance loss, and the consistency of aspect ratio loss to account for the three geometric factors of a bounding box: overlap area, distance, and aspect ratio. It can be defined as:

\begin{equation}
\mathcal{L}_{CIoU} = \mathcal{L}_{IoU} + \mathcal{L}_{NCent} + \alpha \mathcal{L}_{Aspect},
\end{equation}
where
\begin{equation}
IoU = \frac{B \cap B^{gt}}{B \cup B^{gt}}, \quad \mathcal{L}_{IoU} = 1 - IoU,
\label{eq:IoU_loss}
\end{equation}
% \begin{equation}
% \mathcal{L}_{IoU} = 1 - IoU,
% \end{equation}
\begin{equation}
\mathcal{L}_{Cent} = ||p - p^{gt}||^2, \quad \mathcal{L}_{NCent} = \frac{\mathcal{L}_{Cent}}{c^2},
\label{eq:Cent_loss}
\end{equation}
\begin{equation}
\mathcal{L}_{Aspect} = \frac{4}{\pi^2} \left( \arctan \frac{w^{gt}}{h^{gt}} - \arctan \frac{w}{h} \right)^2,
\end{equation}
\begin{equation}
\alpha = 
\begin{cases} 
0, & \text{if } IoU < 0.5, \\
\frac{\mathcal{L}_{Aspect}}{\mathcal{L}_{IoU} + \mathcal{L}_{Aspect}}, & \text{if } IoU \geq 0.5,
\end{cases}
\end{equation}
and \( p \) and \( p^{gt} \) are the central points of the predicted box \( B \) and ground truth box \( B^{gt} \), respectively. \( c \) is the diagonal length of the smallest enclosing box covering the two boxes. \( w \) and \( h \) are the width and height of the box. 
Additionally, we employ the Distribution Focal Loss (DFL) \cite{li2020generalized} strategy to address the problem of the location uncertainty of bounding boxes. Unlike traditional methods that directly regress the vertices of the bounding box, DFL predicts a probability distribution \( P(y) \) to derive the bounding box coordinates as \(\sum_{i=0}^{n} P(y_i) y_i\). The DFL is defined as follows:
\begin{equation}
\begin{aligned}
\mathcal{L}_{DFL} = -((y_{i+1} - y) \log(P(y_i))\ + \\
(y - y_i) \log(P(y_{i+1}))),
\end{aligned}
\end{equation}
where \( y \) is ground truth label, \( y_i \) and \( y_{i+1} \) are the two integers closest to \( y \).

\begin{figure}[htbp] 
	\centering
	\includegraphics[width=0.45\textwidth]{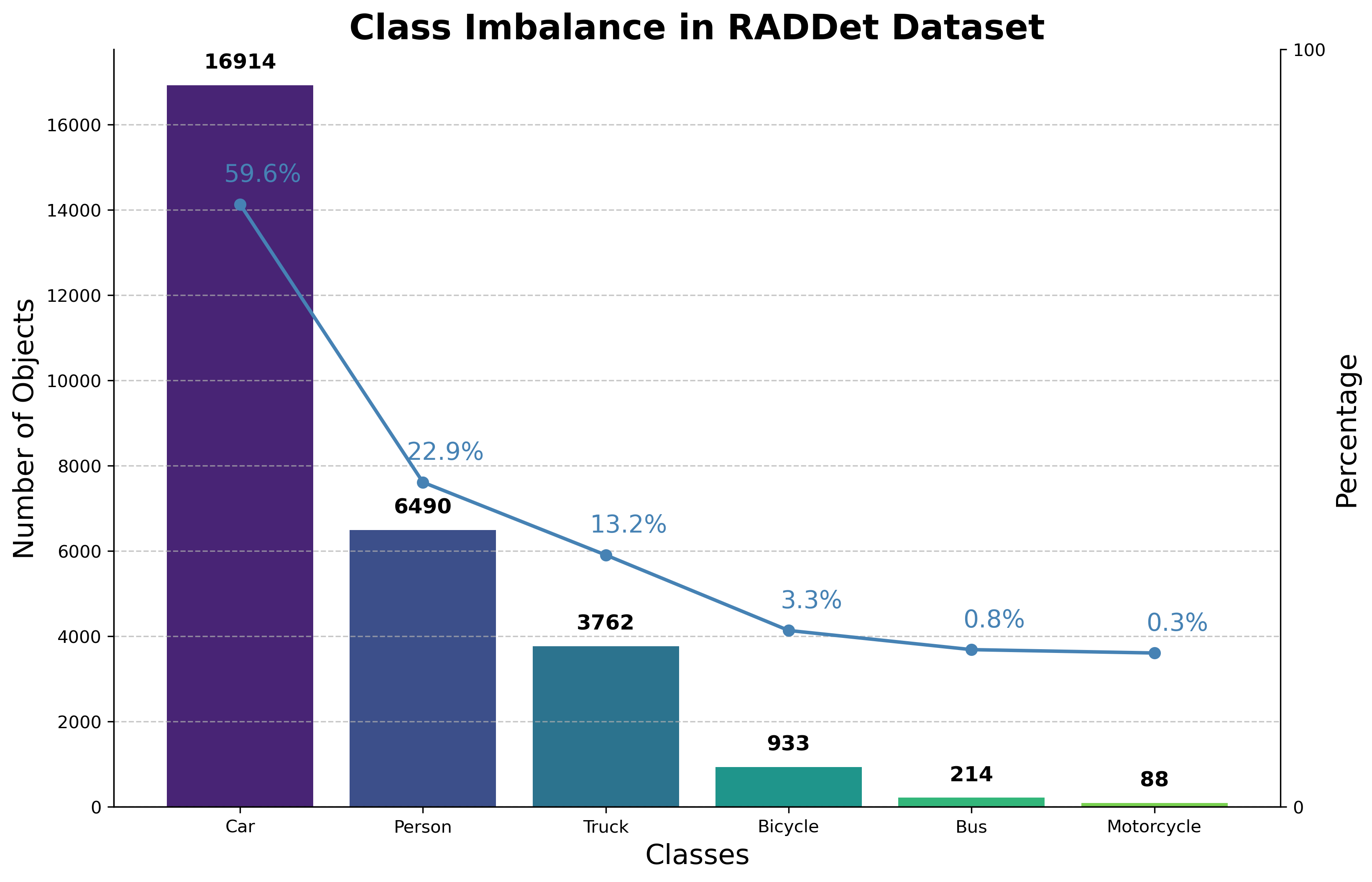}
	\caption{Class imbalance in RADDet dataset.}
	\label{class}
\end{figure}

Second, for the objectness prediction and classification tasks, we choose to use Focal Loss \cite{lin2017focal} because it effectively addresses the imbalance between positive and negative classes. The Focal Loss can be expressed in terms of Binary Cross Entropy (BCE) loss as follows:
\begin{equation}
\mathcal{L}_{FL} = \alpha_t (1 - p_t)^\gamma \mathcal{L}_{BCE},
\end{equation}
where
\begin{equation}
\alpha_t = \alpha * y + (1 - \alpha) * (1 - y), 
\end{equation}
\begin{equation}
p_t = p * y + (1 - p) * (1 - y), 
\end{equation}
\begin{equation}
\mathcal{L}_{BCE} = -w \left[ y \cdot \log p + (1 - y) \cdot \log (1 - p) \right],
\end{equation}
and \( p \) denotes the predicted probability for the positive class, \( \alpha \) is the weighting factor to balance positive and negative classes, \( \gamma \) is the modulating factor to balance easy and hard samples and \( w \) is a weight given to the loss of each class. In our case, we use the default settings of \( \alpha = 0.25 \) and \( \gamma = 2 \). For the weight \( w \), it is set to 1 for objectness prediction. However, for the classification task, since the RADDet \cite{zhang2021raddet} dataset we use suffers from severe class imbalance, as shown in Fig. \ref{class}, with certain classes (such as 'motorcycle', 'bus', and 'bicycle') being significantly underrepresented, it is necessary to apply different weights to balance the loss contributed by different classes.
We propose to calculate the class weights \( \mathbf{w} \) for a batch of samples by first computing the initial class weights based on the inverse frequency of each class: 
\begin{equation}
\mathbf{w} = \frac{(\sum_{j=1}^{C} N_j) - \mathbf{N}}{\sum_{i=1}^{C} ((\sum_{j=1}^{C} N_j) - N_i)},
\end{equation}
where \( \mathbf{N} = [N_1, N_2, \ldots, N_C] \) is the vector of class counts, \( N_i \) is the count of samples in class \( i \), and \( C \) is the total number of classes.
Next, we adjust the weights to ensure a minimum weight \( w_{\text{min}} \) (\(\ 0.05 \) in our case) and normalize them:
\begin{equation}
\mathbf{w} = \max(\mathbf{w}, w_{\text{min}}), \quad \mathbf{w} = \frac{\mathbf{w}}{\sum_{i=1}^{C} w_i}.
\end{equation}

Third, for the Doppler-specific regression task, we choose to use the Smooth L1 loss \cite{girshick2015fast} because it is less sensitive to outliers and better handles unbounded regression targets. This aligns perfectly with our task of predicting the maximum and minimum points in the Doppler dimension. The formula for Smooth L1 loss is given by:
\begin{equation}
\mathcal{L}_{SL1} = 
\begin{cases} 
0.5 (p - y)^2 , & \text{if } |p - y| < 1 \\
|p - y| - 0.5, & \text{otherwise}
\end{cases}
\end{equation}

Fourth, we introduce a 2D box center loss to account for the importance of object centers in radar detection. This loss also helps to improve bounding box localization accuracy, reduce bounding box prediction jitter, and accelerate convergence. To avoid extra computation, we directly use the \( \mathcal{L}_{Cent} \) (see Eq. \ref{eq:Cent_loss}) term already computed in the CIoU loss as our center loss.

Finally, our total loss is the weighted sum of these individual losses, as defined by the following equation:
\begin{equation}
\begin{aligned}
\mathcal{L}_{Total} = &\alpha_1 \mathcal{L}_{FL}^{Obj} + \alpha_2 \mathcal{L}_{FL}^{Cls} + \\ 
&(\alpha_3 \mathcal{L}_{CIoU}^{RA} + \alpha_4 \mathcal{L}_{Cent}^{RA} + 
\alpha_5 \mathcal{L}_{DFL}^{RA})t + 
\\
&\alpha_6 \mathcal{L}_{CIoU}^{RD} + \alpha_7 \mathcal{L}_{Cent}^{RD} + 
\\
&\alpha_8 \mathcal{L}_{SL1}^{Dopl} + \alpha_9 \mathcal{L}_{IoU}^{RAD},
\end{aligned}
\end{equation}
where the weights \(\alpha_1, \alpha_2, \ldots, \alpha_9\) are applied to balance the contributions of the individual losses. Based on our experiments, we found that setting these weights \( \bm{\alpha} = [ 30, 7.5, 7.5, 0.5, 1.5, 5.0, 5.0, 80, 40 ]\), respectively, is a straightforward and reasonable choice. These particular values ensure that each individual loss term remains comparable during the initial training epochs, preventing any one or several loss components from dominating the overall loss value in the training process. Consequently, this approach helps improve both the overall performance and convergence of the model.
The proposed loss function integrates 2D bounding box and center losses for the radar's RA and RD planes, to ensure robust 2D detection capabilities and reflect the common practice of presenting radar detections in 2D bird's-eye views.
To reduce computational overhead, we only compute the DFL loss on the RA plane. Furthermore, since we have already incorporated the losses from both RA and RD planes, a simple IOU loss calculation (see Eq. \ref{eq:IoU_loss}) for the 3D bounding box is sufficient.
Importantly, the loss calculation for the RA plane is re-weighted using the anchor alignment metric \( t \), as we apply TAL only for label assignment in the RA plane. The original TAL method employs a normalized anchor alignment metric \(\hat{t}\), which normalizes to make the maximum of \(\hat{t}\) equal to the largest IoU value within each instance \cite{feng2021tood}. However, we observed that using \(\hat{t}\) hinders our model's convergence. This issue likely arises due to the high noise in radar data and the small size of radar targets, which result in generally much lower IoU scores. Consequently, \(\hat{t}\) becomes too small, failing to provide an effective supervisory signal to encourage the model to continue learning. Therefore, we opt to use the original anchor alignment \( t \) metric in our loss function. It is worth noting that after model convergence, adjusting \(\alpha_1 = 40\) and \(\alpha_2 = 15\) for a second round of transfer learning can significantly improve objectness and classification accuracy without impairing localization accuracy.

\subsection{Location-Aware NMS}
\label{sec:nms}
Object detection typically yields multiple bounding boxes for a single object, which are then filtered out through a post-processing step involving NMS\cite{symeonidis2023neural}. NMS suppresses duplicate detections by selecting the bounding boxes with the highest classification scores among those belonging to the same class. However, this process fails to address the spatially overlapping bounding boxes belonging to different classes, resulting in reduced performance. Moreover, the effectiveness of NMS is heavily reliant on classification accuracy, which is a known weakness of radar data that can lead to excessive duplicate radar detections from different classes.

To mitigate this limitation, we propose Location-Aware NMS, which further refines the filtering process by considering the spatial overlap between bounding boxes belonging to different classes.  Specifically, Location-Aware NMS calculates the IoU between bounding boxes of different classes and removes the one with a lower classification score if the IoU exceeds a predefined threshold (set to 0.1 in our case), as demonstrated in Algorithm \ref{alg1}. This approach effectively avoids the occurrence of multiple bounding boxes corresponding to the same object but belonging to different classes, a scenario that traditional NMS fails to effectively prevent. Our proposal of Location-Aware NMS is grounded in two key facts: (1) radar object detection exhibits higher location accuracy than classification accuracy, and (2) the probability of radar targets overlapping in RAD data is relatively low. 

\renewcommand{\algorithmiccomment}[1]{\hfill$\triangleright$\textit{\textcolor{blue}{#1}}}
\begin{algorithm}
\caption{Location-Aware NMS (LA-NMS)}\label{alg1}
\textbf{Input: }$\text{bbox} \in \mathbb{R}^{N \times 8}$, \ $\text{thr} \in \mathbb{R}$\\
\textbf{Output: }$\text{selected\_bbox} \in \mathbb{R}^{M \times 8}$ \\
\textbf{Note:} bbox format: $[x, y, z, w, h, d, \text{class\_score}, \text{class}]$
\begin{algorithmic}[1]
\State $\text{bbox} \gets \text{bbox}[\text{argsort}(-\text{bbox}[:, 6])]$ \Comment{Sort by class\_score}
\State $\text{selected\_bbox} \gets [\ ]$ 

\While {$\text{len}(\text{bbox}) > 0$}
    \State $\text{current\_box} \gets \text{bbox}[0]$ \Comment{Select highest score box}
    \State $\text{selected\_bbox.append}(\text{current\_box})$
    \State $\text{bbox} \gets \text{bbox}[1:]$ \Comment{Remove selected box}
    \State $\text{ious} \gets [\ ]$
    \For {\text{box in bbox}}   \Comment{Compute IoU}
        \State $\text{iou} \gets \text{compute\_iou}(\text{current\_box}, \text{box})$
        \State $\text{ious.append(iou)}$
    \EndFor
    
    \State $\text{keep\_indices} \gets [\ ]$ \Comment{Check IoU and class}
    \For {\text{i, box in enumerate(bbox)}} 
        \If {$\text{ious}[i] \leq \text{thr} \text{ or } \text{box}[7] == \text{current\_box}[7]$}
            \State $\text{keep\_indices.append(i)}$
        \EndIf
    \EndFor
        
    \State $\text{bbox} \gets \text{bbox}[\text{keep\_indices}]$ \Comment{Keep remaining boxes}
\EndWhile \\
\Return $\text{selected\_bbox}$
\end{algorithmic}
\end{algorithm}

\section{Experiments}
\subsection{Dataset}
Our model was evaluated on the RADDet dataset \cite{zhang2021raddet}, which, to the best of our knowledge, is currently the only real-world radar raw data dataset providing RAD data with 3D bounding box annotations across the range, azimuth, and Doppler dimensions. Other datasets either only supply radar point cloud data, like NuScenes and Waymo, or only provide 2D bounding box annotations, like CARRADA and RADIATE, or restrict to point annotations, like CRUW \cite{paek2022k}. 
The recently introduced K-radar \cite{paek2022k} dataset provides 4D radar data (incorporating Doppler, range, azimuth, and elevation) along with 3D bounding box annotations. However, these 3D annotations only cover the range, azimuth, and elevation dimensions—lacking the Doppler dimension—while our study prioritizes 3D detection across the range, azimuth, and Doppler dimensions for determining distance, angle, and speed.

The RADDet dataset was collected using a Texas Instruments AWR1843-BOOST radar and a pair of DFK 33UX273 stereo cameras in real road environments under sunny conditions. The radar sensor consists of 2 transmitters and 4 receivers, producing 64 chirps each sampled 256 times, resulting in raw ADC data with a shape of (256, 8, 64). This ADC data undergoes 3D-FFT processing, resulting in a RAD cube frame of shape (256, 256, 64). The dataset comprises 10,158 frames, containing 28,401 objects classified into six categories: "person", "bicycle", "car", "motorcycle", "bus", and "truck". Each object is annotated with a 3D bounding box, formatted as \( [ x_{center},y_{center},z_{center}, w,h, d ] \).
Our model uses the 3D RAD data as input. To facilitate model processing, we resized the original RAD data from its shape of (256, 256, 64) to (256, 256, 256) by performing nearest neighbor interpolation \cite{Interpolation} along the Doppler dimension, resulting in uniform dimensions across all axes. The nearest neighbor interpolation was chosen for two reasons: it fills gaps with the nearest data points, preserving the original data distribution, which is critical for radar data analysis, and it maintains the shape and center of radar objects, allowing for direct use of the original annotations after scaling.

\begin{figure*}[htpb]
	\centering
	\includegraphics[width=0.999\textwidth]{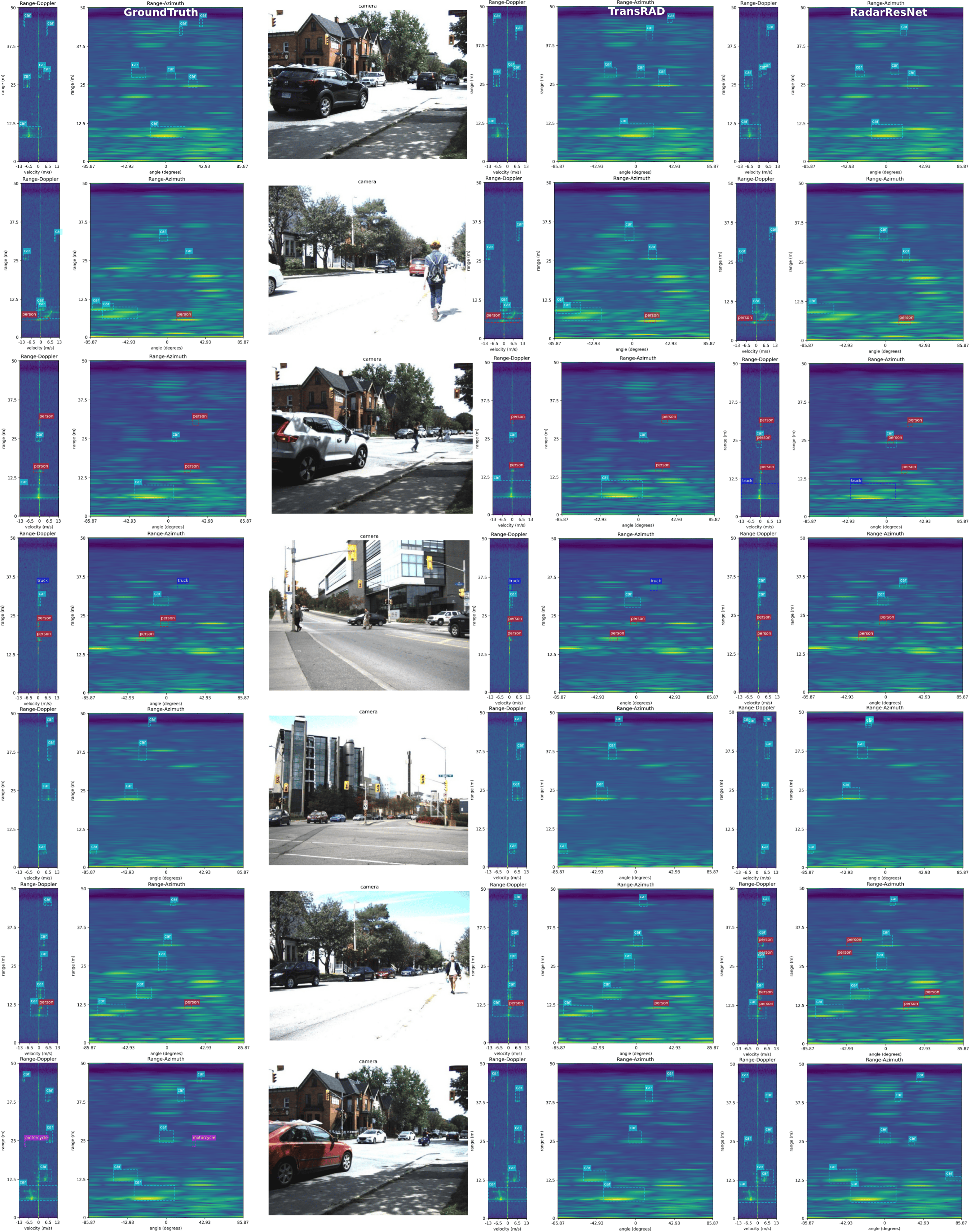}
	\caption{Radar object detection results comparison between the ground truth, TransRAD, and RadarResNet.}
	\label{det_result}
\end{figure*}

\subsection{Evaluation Metrics}
The widely used evaluation metrics, mAP, FLOPs, model parameters, and inference time, are employed to assess the performance of our detector. \subsubsection{mAP} The mean Average Precision (mAP) is determined by calculating the Average Precision (AP) for each class and averaging these values across all classes under a specific IoU threshold. AP is designed to comprehensively evaluate model performance by considering both the precision (\(P\)) and recall (\(R\)) of the detector \cite{padilla2020survey}. As shown in the equations below, AP is high when both precision and recall are high, and low when either is low across a range of threshold values.
\begin{equation}
\small
\text{AP} = \int_{0}^{1} P(R) \, dR \approx
\sum_{n} (R_{n+1} - R_n) \max(P_{n+1} , P_n),
\end{equation} %\equiv 
where
\begin{equation}
\small
\text{P} = \frac{\text{TP}}{\text{TP} + \text{FP}}, \quad \text{R} = \frac{\text{TP}}{\text{TP} + \text{FN}},
\end{equation}
and \(P_n\) and \(R_n\) are the precision and recall at the \(n\)th threshold, TP is True positive, FP is False positive, and FN is False negative. The mAP is the average AP over all classes and over multiple IoU thresholds, that is
\begin{equation}
\text{mAP} = \frac{1}{T} \sum_{t=1}^{T} \frac{1}{N} \sum_{n=1}^{N} \text{AP}_t(\text{IoU}_n),
\end{equation}
where \(\text{AP}_t(\text{IoU}_n)\) is the AP for the \(t\)th class at the \(n\)th IoU threshold, \(T\) is the total number of classes, and \(N\) is the total number of IoU thresholds being evaluated. 
In our experiments, we use AP metrics ranging from 0.5 to 0.9 for 2D detection and from 0.3 to 0.7 for 3D detection, both with a step size of 0.1.

\subsubsection{FLOPs} FLOPs (Floating Point Operations) represent the number of arithmetic operations—addition, subtraction, multiplication, and division—performed on floating-point numbers. They are fundamental in machine learning computations, including matrix multiplications, activations, and gradient calculations, and serve as a key metric for assessing a model's computational cost and efficiency \cite{vu2020not}. In our experiments, we use GFLOPs (Giga Floating Point Operations) to evaluate models' computational complexity.

\subsubsection{Model Parameters} Model parameters, specifically weights and biases in deep learning, are the training parameters that are learned during the learning process. This metric reflects the capacity and potential complexity of the model. In particular, a higher number of parameters can capture more intricate features but may increase the risk of overfitting and demand greater memory and computational resources. Conversely, a model with fewer parameters tends to be simpler, faster, and less prone to overfitting but may struggle to learn meaningful features. We use millions of parameters as a unit to evaluate model complexity.

\subsubsection{Inference Time} Inference time refers to the duration required for a trained model to make predictions on new, unseen data. This metric is crucial for evaluating the model's efficiency and real-time applicability. Faster inference times are often desired in practical applications where rapid decision-making is essential, such as autonomous driving or real-time object detection. However, achieving lower inference times may involve trade-offs, such as reducing model complexity, which can impact accuracy. We estimate inference time by measuring the time taken for the model to detect a single frame, expressed in milliseconds per frame.

\subsection{Training Details}
All experiments were performed on a single Nvidia 32GB V100S GPU and an AMD Zen2 processor with 5 cores and 30GB of RAM, using PyTorch 2.0 as the deep learning framework. All models were trained from scratch and used the same hyperparameters. Specifically, we employed the Adam optimizer with a momentum of 0.937 and a weight decay of 0. A cosine learning rate schedule was used to adjust the learning rate, with a warmup ratio of 0.05, an initial learning rate of 0.001, and a minimum learning rate of 0.00001. The total number of epochs was set to 100, with a batch size of 4. Additionally, the exponential moving average (EMA) strategy with a decay rate of 0.9999 was adopted during the training process. No data augmentation techniques were used except for resizing the data.

\subsection{Comparisons With SOTAs}

\begin{table*}[ht]
    \centering
    \caption{3D DETECTION PERFORMANCE COMPARISON OF DIFFERENT METHODS}
    \setlength{\tabcolsep}{4pt} % Adjust column separation
    \renewcommand{\arraystretch}{1.2} % Adjust row separation
    \begin{adjustbox}{width=\textwidth}
    \begin{tabular}{lccccccccc}
        \toprule
        \multicolumn{1}{l}{\multirow{2}{*}{Method}} & \multicolumn{1}{c}{Params\(\downarrow\)} & \multicolumn{1}{c}{FLOPs\(\downarrow\)} & \multicolumn{1}{c}{Inference Time\(\downarrow\)} & \multicolumn{5}{c}{mAP(\(\%\))\(\uparrow\) for RAD 3D Detection} \\
        \cmidrule(lr){2-2} \cmidrule(lr){3-3} \cmidrule(lr){4-4} \cmidrule(lr){5-9}
        & (M) & (G) & (ms per frame) & $AP_{0.3}$ & $AP_{0.4}$ & $AP_{0.5}$ & $AP_{0.6}$ & $AP_{0.7}$ \\
        \midrule
        \rowcolor{gray!10}RadarResNet\cite{zhang2021raddet}  & 8.07 & 4.86 & 16.57 & 52.90 & 34.83 & 19.41 & 8.66 & 2.78  \\
        RODNET-CDC\cite{wang2021rodnet2}  & 36.89 & 172 & 21.40 & 51.12 & 37.74 & 24.53 & 12.71 & 5.30   \\
        RODNET-HG\cite{wang2021rodnet2} & 132 & 178.4 & 55.78 & 37.14 & 26.47 & 15.85 & 6.80 & 2.29   \\
        RAMP-CNN\cite{gao2020ramp} & 107 & 487 & 15.81 & 44.04 & 31.63 & 19.29 &9.39 & 4.16   \\
        T-RODNet\cite{jiang2022t}  & 161 & 175 & 19.17 & 22.57 & 15.91 & 10.81 & 6.62 & 3.37   \\
        RadarFormer\cite{dalbah2023radarformer} & 6.19 & 37.47 & 12.88 & 38.09 & 22.80 & 14.33 & 8.05 & 3.83   \\
        YOLOv8\cite{Yolov8} & 18.14 & \textbf{1.36} & 5.49 & 42.01 & 31.27 & 20.65 & 10.48 & 4.28   \\
        \midrule
         \rowcolor{gray!8}TransRAD & \textbf{5.78} & 2.16 & \textbf{4.37} & \textbf{61.90} & \textbf{50.89} & \textbf{38.76} & \textbf{25.54} & \textbf{13.17}  \\
        \bottomrule %\rowcolor{gray!20}
    \end{tabular}
    \end{adjustbox}    
    \label{tab:3D performance_comparison}
\end{table*}

\begin{table*}[ht]
    \centering
    \caption{2D DETECTION PERFORMANCE COMPARISON OF DIFFERENT METHODS}
    \setlength{\tabcolsep}{4pt} % Adjust column separation
    \renewcommand{\arraystretch}{1.2} % Adjust row separation
    \begin{adjustbox}{width=\textwidth}
    \begin{tabular}{lccccccccccc}
        \toprule
        \multicolumn{1}{l}{\multirow{2}{*}{Method}} & \multicolumn{5}{c}{mAP(\(\%\))\(\uparrow\) for RA 2D Detection} & \multicolumn{5}{c}{mAP(\(\%\))\(\uparrow\) for RD 2D Detection} \\
        \cmidrule(lr){2-6} \cmidrule(lr){7-11}
        & $AP_{0.5}$ & $AP_{0.6}$ & $AP_{0.7}$ & $AP_{0.8}$ & $AP_{0.9}$ & $AP_{0.5}$ & $AP_{0.6}$ & $AP_{0.7}$ & $AP_{0.8}$ & $AP_{0.9}$ \\
        \midrule
        \rowcolor{gray!10}RadarResNet\cite{zhang2021raddet}  & 45.35 & 29.29 & 14.78 & 4.05 & 0.29 & 39.45 & 24.73 & 13.33 & 4.19 & 0.49  \\
        RODNET-CDC\cite{wang2021rodnet2}  & 46.38 & 34.52 & 19.88 & 6.37 & 0.63 & 39.23 & 26.61 & 15.97 & 6.78 & 1.59   \\
        RODNET-HG\cite{wang2021rodnet2} & 33.14 & 22.15 & 11.12 & 3.24 & 0.38 & 29.19 & 17.77 & 8.65 & 2.56 & 0.11   \\
        RAMP-CNN\cite{gao2020ramp} & 39.72 & 27.33 & 14.56 & 4.48 & 0.38 & 33.60 & 20.86 & 10.95 & 4.75 & 0.75   \\
        T-RODNet\cite{jiang2022t}  & 18.98 & 15.36 & 11.69 & 3.43 & 0.13  & 21.98 & 13.90 & 6.25 & 2.75 & 0.75  \\
        RadarFormer\cite{dalbah2023radarformer} & 31.38 & 23.82 & 16.25 & 4.18 & 0.20 & 28.30 & 13.52 & 7.86 & 3.49 & 0.96   \\
        YOLOv8\cite{Yolov8} & 37.03 & 25.86 & 13.08 & 3.65 & 0.17  & 33.88 & 23.52 & 13.46 & 5.87 & 0.92  \\
        DAROD\cite{decourt2022darod} & 32.14 & 21.14 & 10.86 & 1.78 & 0.02  & 44.19 & 30.35 & 16.08 & 4.04 & 0.11  \\        
        \midrule
         \rowcolor{gray!8}TransRAD & \textbf{55.90} & \textbf{45.78} & \textbf{32.16} & \textbf{14.27} & \textbf{1.64} & \textbf{51.80} & \textbf{40.27} & \textbf{27.37} & \textbf{14.16} & \textbf{3.63}  \\
        \bottomrule %\rowcolor{gray!20}
    \end{tabular}
    \end{adjustbox}    
    \label{tab:2D performance_comparison}
\end{table*}

\newcommand{\fullcircle}{\ding{108}} % Full circle
\newcommand{\emptycircle}{\ding{109}} % Empty circle
\begin{table*}[h!] %The Effect of Each Module on the TRANSRAD Model
    \centering
    \caption{ABLATION RESULTS OF TRANSRAD}
    \begin{adjustbox}{width=\textwidth}
    \begin{tabular}{ccccccccc}
        \toprule
        \multicolumn{1}{l}{\multirow{2}{*}{FPN Neck}} & \multicolumn{1}{l}{\multirow{2}{*}{Decoupled Head}} & \multicolumn{1}{l}{\multirow{2}{*}{2D and Center Loss}} & \multicolumn{1}{l}{\multirow{2}{*}{LA-NMS}} & \multicolumn{5}{c}{mAP(\(\%\))\(\uparrow\) for RAD 3D Detection} \\     \cmidrule(lr){5-9}
        & & & &$AP_{0.3}$ &$AP_{0.4}$ &$AP_{0.5}$ &$AP_{0.6}$ & $AP_{0.7}$ \\
        \midrule
        \rowcolor{gray!10} \fullcircle & \fullcircle & \fullcircle & \fullcircle & {61.90} & {50.89} & {38.76} & {25.54} & {13.17} \\
        \emptycircle & \fullcircle & \fullcircle & \fullcircle & 55.51 & 46.16 & 34.21 & 21.77 & 11.14 \\
        \fullcircle & \emptycircle & \fullcircle & \fullcircle & 59.75 & 49.21 & 38.03& 24.82 & 13.18 \\
        \fullcircle & \fullcircle & \emptycircle & \fullcircle & 59.76 & 49.12 & 37.00& 24.81 & 12.88 \\
        \fullcircle & \fullcircle & \fullcircle & \emptycircle & 59.14 & 48.44 & 35.47& 22.42 & 11.51 \\
        \bottomrule
    \end{tabular}  
    \end{adjustbox}
    \label{tab:ablation_study}
\end{table*}

To assess the performance of our proposed approach, we conducted a comprehensive comparison with several SOTA methods, with the results presented in Tables \ref{tab:3D performance_comparison} and \ref{tab:2D performance_comparison}. In this comparison, RadarResNet serves as the baseline, and both RadarResNet and DAROD are used without any modifications. The remaining models only utilize their Backbone parts, sharing the same Neck, Head, and loss functions as our proposed method.
\subsubsection{Quantitative Performance Comparison} 
Table \ref{tab:3D performance_comparison} presents the quantitative comparisons of our proposed approach with some SOTA methods for 3D radar object detection. The experimental results indicate that our TransRadar outperforms SOTA methods in almost all metrics except for FLOPS. Our model has the smallest model parameter count compared to other models, attributed to our lightweight Backbone. As discussed in Sec. \ref{sec:backbone}, radar data is not as complex, thus a lightweight model suffices. Moreover, our model has the second-lowest FLOPS, with a value only slightly higher than YOLOv8 and significantly lower than all other models. This is mainly due to the fact that most SOTA models, except for RadarResNet, employ 3D CNN, a highly computationally intensive network, and tend to use much deeper networks, ignoring the design of efficient lightweight networks tailored to radar data features. RadarResNet’s slightly higher FLOPs are due to its 3D grid-based bounding box regression, which increases its computational cost. Furthermore, our model achieves the fastest inference and highest accuracy. Specifically, our model's inference speed is significantly faster (3-10 times) than all other models except YOLOv8 (slightly faster than it), which is currently one of the fastest real-time image detectors. In terms of object detection accuracy, our model shows significant improvements in 3D detection over all other models. Compared to the baseline (RadarResNet), our model achieves a 9\% improvement in $AP_{0.3}$ and a 16\% improvement in $AP_{0.4}$, with $2\times$, $3\times$, and $5\times$ improvements in $AP_{0.5}$, $AP_{0.6}$, and $AP_{0.7}$, respectively. Compared to RODNet-CDC, the best-performing SOTA model in this experiment, our model improves by 10\% in $AP_{0.3}$, 13\% in $AP_{0.4}$, and $1.5\times$, $2\times$, and $2.5\times$ in $AP_{0.5}$, $AP_{0.6}$, and $AP_{0.7}$, respectively. As for the remaining models, it shows at least $1.5\times$ and up to $6\times$ improvement. Similar performance enhancements can be observed in Table \ref{tab:2D performance_comparison} for 2D RA and RD detection. This strongly validates the effectiveness of our TransRAD model. Additionally, we observe that the higher the $AP$ threshold, the greater the performance lead of our model over others, indicating that our model can more accurately locate targets. This may be attributed to our RMT-based Backbone, which explicitly incorporates spatial priors to better match the radar targets. In contrast, other models, despite having more parameters and complex network structures, fail to achieve desirable results. This is likely because they are directly adapted from image object detection models without considering the unique aspects of radar object detection, such as low resolution, high noise, and lack of appearance information (more discussed in Sec. \ref{sec:Radar Object Detection}). Our implementation of the YOLOv8 radar detector also demonstrates this point, achieving an $AP_{0.3}$ accuracy of only 42.01\%. Another reason may be that our radar dataset is relatively small, making it challenging for the heavy models to learn effectively. Notably, the two transformer-based radar detectors, T-RODNet and RadarFormer, exhibit the poorest performance among the SOTA models. The analysis of the training process reveals that these models struggle to converge on the training dataset, indicating that they cannot effectively learn useful features from the high-noise and class-imbalanced RADDet radar dataset. This may be attributed to the fact that plain transformer models are less effective in handling high-noise spatial data and are more sensitive to limited datasets compared to CNN models.
Finally, Table \ref{tab:2D performance_comparison} provides the results of our model on 2D RA and RD detection, demonstrating significant improvements over other models. This may be attributed to the integration of 2D detection loss in our model.
\begin{figure}[htpb]
	\centering
	\includegraphics[width=0.48\textwidth]{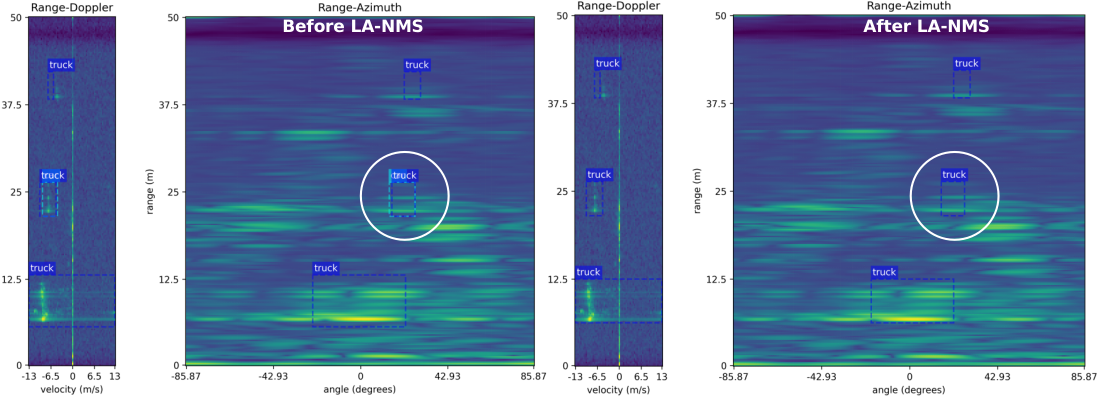}
	\caption{LA-NMS removes overlapping bounding boxes of different classes for the same object.}
	\label{nms_result}
\end{figure}

\subsubsection{Qualitative Performance Comparison}

Fig. \ref{det_result} provides a visual comparison between the detection results of our TransRAD, the ground truth, and the baseline, demonstrating the reliability and accuracy of our model. On one hand, TransRAD's results closely match the ground truth in both classification and localization accuracy. On the other hand, TransRAD significantly surpasses the baseline in both aspects.
Firstly, the baseline misses detecting a small and a large target, as shown in the first and second rows of Fig. \ref{det_result}, respectively. This is primarily attributed to the RadarResNet's lack of specialized adjustments for radar small target detection and its failure to utilize multi-scale feature fusion structures similar to FPN for handling different object sizes.
Secondly, the baseline, as demonstrated in the third and fourth rows of Fig. \ref{det_result}, misclassifies the "car" and "truck" classes. In contrast, our model accurately classifies these classes, highlighting its superior classification performance. 
Thirdly, the third row of Fig. \ref{det_result} illustrates the issue of two overlapping bounding boxes corresponding to the same target but belonging to different classes, which is not present in our model. This underscores the effectiveness of our proposed LA-NMS in mitigating this problem. 
Fourthly, the fifth and sixth rows of Fig. \ref{det_result} showcase some false detections by the baseline, most likely due to inaccurate objectness predictions. In contrast, our model alleviates this issue using a dedicated objectness detection head. 
The aforementioned results collectively demonstrate that our TransRAD model has achieved a notable performance improvement compared to the baseline.
Nevertheless, our model is not perfect, as illustrated by the final row of Fig. \ref{det_result}, where it failed to detect a motorcycle, whereas the baseline, although misclassifying it, successfully detected the target.

\subsection{Ablation Study}

To verify the respective effectiveness of each key module in our proposed TransRAD radar detector, we conduct an ablation study to assess their individual effects on the overall performance. The ablation results are provided in Table \ref{tab:ablation_study}. Notably, Tables \ref{tab:3D performance_comparison} and \ref{tab:2D performance_comparison} have already validated the effectiveness of our RMT-based Backbone, so an ablation study on the Backbone is unnecessary.

\subsubsection{FPN Neck}
We replaced the FPN Neck with a Neck that produces a single-scale feature map and performed object detection based on this feature map. As shown in the second row of Table \ref{tab:ablation_study}, detection performance significantly declined, with $AP_{0.3}$ dropping from 61.90\% to 55.51\%. This highlights the importance of the FPN in radar object detection. As discussed in Sec. \ref{sec:neck}, the FPN generates three feature maps at different scales and effectively integrates high-level semantic features with low-level detailed features, which is crucial for detecting small radar targets.
\subsubsection{Decoupled Head}
We replace our four independent decoupled heads with a single coupled head, which simultaneously predicts four different tasks. As shown in the third row of Table \ref{tab:ablation_study}, the detection performance experiences a certain decline, with $AP_{0.3}$ accuracy decreasing from 61.90\% to 59.75\%. This is anticipated, as discussed in Sec. \ref{sec:head}, that using a single head with the same parameters to handle distinctly different tasks will inevitably cause conflicts, leading to suboptimal performance for each task. Decoupled heads, by using dedicated heads for each task, can alleviate this issue. In particular, dedicated classification and objectness heads can significantly alleviate the issues frequently encountered by deep radar detectors, as noted in \cite{zhang2021raddet}, namely misclassification and false positive detection.
\subsubsection{2D Bounding Box Loss and Center Loss}
When we remove the Center Loss and 2D Bounding Box Loss for RA and RD planes from TransRAD, we observe that the model converges slower and the performance declines, as shown in the fourth row of Table \ref{tab:ablation_study}. Specifically, $AP_{0.3}$ accuracy decreases from 61.90\% to 59.76\%, with greater declines at higher AP thresholds, indicating reduced localization accuracy. Therefore, the 2D Bounding Box Loss and Center Loss are essential for more accurate bounding box regression. Additionally, the radar 2D detection results presented in Table \ref{tab:2D performance_comparison} also demonstrate our model's superior performance on RA and RD 2D planes, which is partly attributed to the adoption of explicit 2D Bounding Box Loss. In fact, radar object detection prefers to obtain more accurate center positioning and 2D plane detection results, making the integration of 2D Bounding Box Loss and Center Loss necessary.
\subsubsection{Location-Aware NMS}
Disabling LA-NMS led to a noticeable performance drop, with $AP_{0.3}$ falling from 61.90\% to 59.14\%, as shown in the fifth row of Table \ref{tab:ablation_study}. In fact, radar detectors tend to produce duplicate bounding boxes for the same object, and traditional NMS can only remove duplicate bounding boxes belonging to the same class, failing to suppress those of different classes. Our proposed LA-NMS can address this issue, as illustrated in Fig. \ref{nms_result}. Moreover, the probability of overlapping radar objects in RAD data is extremely low, further justifying the use of LA-NMS. It is particularly noteworthy that we found achieving 3D radar detection to be essential. Otherwise, a large number of overlapping 2D bounding boxes would appear on both the 2D RA plane and RD plane, despite not overlapping in 3D space. Consequently, using NMS would inevitably result in the erroneous removal of bounding boxes that contain valid targets.

\section{Conclusions}
In this paper, we proposed TransRAD, a robust radar object detection model that effectively addresses the unique challenges posed by radar data. Our model integrates an RMT Backbone, an FPN Neck, an anchor-free decoupled Head, and a Location-Aware Non-Maximum Suppression (LA-NMS) technique, aimed at performing 3D object detection on radar RAD data. The experimental results on the RADDet dataset highlight the superiority of TransRAD over existing SOTA methods in terms of detection accuracy, inference speed, and computational efficiency. Specifically, the use of the RMT Backbone with explicit spatial priors enabled our model to learn truly useful features from challenging radar data, thereby enhancing both 3D and 2D radar object detection precision. Additionally, the ablation studies confirmed the critical contributions and effectiveness of each component in enhancing TransRAD's overall detection performance. While our model shows promising results, there remain challenges to address, such as the occasional misclassification and missed detection of objects. Future work will focus on further refining the model to enhance its detection and classification capabilities, as well as exploring its applicability in more diverse and complex real-world environments.
%%%%%%%%%%%%%%%%%%%%%%%%%%%%%%%%%%%%%%%%%%%%%

\vspace{0.5cm}

\bibliographystyle{IEEEtran}
{\small
\bibliography{references.bib}
}

%\end{thebibliography}

% \newpage
\vspace{-20pt}

\vfill

\end{document}